\documentclass[nohyperref]{article}

\usepackage{microtype}
\usepackage{graphicx}
\usepackage{subfigure}
\usepackage{booktabs} 

\usepackage{hyperref}

\usepackage[accepted]{icml2023}

\usepackage{amsmath}
\usepackage{amssymb}
\usepackage{mathtools}
\usepackage{amsthm}

\usepackage[capitalize,noabbrev]{cleveref}

\usepackage[version=4]{mhchem}

\theoremstyle{plain}

\theoremstyle{definition}

\theoremstyle{remark}

\usepackage[textsize=tiny]{todonotes}

\icmltitlerunning{Extracting Molecular Properties from Natural Language with Multimodal Contrastive Learning}

\begin{document}

\twocolumn[
\icmltitle{Extracting Molecular Properties from Natural Language with Multimodal Contrastive Learning}



\begin{icmlauthorlist}
\icmlauthor{Romain Lacombe}{yyy}
\icmlauthor{Andrew Gaut}{yyy}
\icmlauthor{Jeff He}{yyy}
\icmlauthor{David Lüdeke}{yyy}
\icmlauthor{Kateryna Pistunova}{yyy}
\end{icmlauthorlist}

\icmlaffiliation{yyy}{Stanford University, Stanford, CA, United States}

\icmlcorrespondingauthor{Romain Lacombe}{rlacombe@stanford.edu}

\icmlkeywords{Machine Learning, ICML}

\vskip 0.3in
]



\printAffiliationsAndNotice{}  

\begin{abstract}
Deep learning in computational biochemistry has traditionally focused on molecular graphs neural representations; however, recent advances in language models highlight how much scientific knowledge is encoded in text. To bridge these two modalities, we investigate how molecular property information can be transferred from natural language to graph representations. We study property prediction performance gains after using contrastive learning to align neural graph representations with representations of textual descriptions of their characteristics. We implement neural relevance scoring strategies to improve text retrieval, introduce a novel chemically-valid molecular graph augmentation strategy inspired by organic reactions, and demonstrate improved performance on downstream \emph{MoleculeNet} property classification tasks. We achieve a +4.26\% AUROC gain versus models pre-trained on the graph modality alone, and a +1.54\% gain compared to the recently proposed molecular graph/text contrastively trained \emph{MoMu} model \cite{su2022}.

\end{abstract}

\section{Introduction}

Deep molecular representation learning models have demonstrated significant potential for important tasks in computational biology and chemistry, such as predicting molecular properties or screening candidates for drug discovery. However, existing AI models typically focus on either graph-based representations, or knowledge extraction from natural language, leaving a gap between these two modalities.

In this work, we investigate whether learning molecular graph representations jointly with textual representations of the corresponding molecule improves those representations. Specifically, we improve on a recently proposed molecular multimodal model (\emph{MoMu}) for contrastive joint text-graph representation learning \cite{su2022} by enhancing the relevance of natural language property descriptions to which we align neural molecular representations. We implement neural relevance based methods to improve text sampling, and introduce a novel, principled approach for chemically-valid graph augmentation which yields promising results.

We hope our improved multimodal pre-training strategy for property prediction, along with experimental results and identified avenues for future work, contribute to the development of more expressive models for computational biochemistry and molecular sciences.

\section{Related Work}

\paragraph{Molecular representation learning.} Molecular representation learning has played a crucial role in recent computational biology advances. Traditional molecular representations, such as SMILES \cite{weininger1988} which represent molecules as strings of atoms, have limited capacity to capture complex molecular structures. Molecular graph representations \cite{duvenaud2015, kearnes2016} have have been shown to better capture the structural and functional properties of molecules. Graph Convolutional Networks (GCNs) \cite{kipf2016}, Graph Attention Networks (GATs) \cite{velickovic2017}, and Graph Isomorophism Networks (GINs) \cite{xu2018powerful} are popular Graph Neural Network (GNNs) architectures which outperform traditional machine learning algorithms in various molecular prediction tasks \cite{gomez2016, gilmer2017, wu2018}.

\paragraph{Language models in chemistry.} The advent of large language models, such as GPT \cite{radford2018}, BERT \cite{devlin2018}, and T5 \cite{raffel2019}, has transformed the field natural language processing. Researchers have started exploring their potential in cheminformatics, leading to the development of models such as ChemBERTa \cite{korolev2020} and MolBERT \cite{napolitano2021}. These models have shown promising results in tasks like reaction prediction, retro-synthesis, and molecular property prediction \cite{kusner2017}---so much so that \citet{White2023} concludes: ``the future of chemistry is language.''  

\paragraph{Contrastive learning} Contrastive learning has been used to learn good pre-trained representations of data unsupervisedly \cite{you2020graph, wang2022molecular}. Contrastive learning can also be used in a multimodal setting to learn joint representations of the image and text modalities \cite{radford2021}. \citet{su2022} use a similar method to jointly train graph and text encoders and call their model \emph{MoMu}. Our work builds on \emph{MoMu} as our baseline model.

\section{Methods}

\subsection{Foundation Model Paradigm: Pretrain \& Finetune}

We approach the task of extracting molecular properties from natural language through the lens of the foundation model paradigm, following a ``pre-train and fine-tune'' strategy \cite{liang2021foundation} presented in figure \ref{fig:contrastive-learning}, with (1) pre-trained text and graph encoder models, (2) aligned through contrastive learning, then (3) evaluated on downstream classification tasks:

\begin{itemize}
\item We use two previously pre-trained encoders: a bidirectional transformer (SciBERT) for text \cite{beltagy2019scibert}, and a pre-trainGraph Isomorphism Network (GIN) \cite{xu2018powerful} for graphs;
\item We align their representations in a joint latent space through contrastive pre-training over graph-text pairs;
\item We then fine-tune the graph encoder on a series of downstream molecular property prediction tasks, and evaluate the quality of our pre-training based on performance on these downstream tasks.
\end{itemize}

\subsection{Multimodal Contrastive Pre-Training}

\subsubsection{Contrastive Learning Strategy}

The core machine learning task in our approach is to learn aligned representations of pairs of molecular graphs and paragraphs of text in natural language describing the properties of that molecule. We use the self-supervised learning technique of contrastive learning, based on a loss function which promotes smaller euclidian distances in the joint latent space between graph and text samples of the same data samples (positive pairs), and larger euclidian distances between non-matching samples (negative pairs). 

Building on the original \emph{MoMu} implementation \cite{su2022}, we use the following contrastive learning paradigm:
\begin{itemize}
    \item Form of batch of $N$ molecules $i \in [1, ..., N]$;
    \item Sample $2 N$ relevant text fragments where $\{\mathcal{T}_i^1, \mathcal{T}_i^2\}$ describe molecule $i$;
    \item From the original $\mathcal{G}_i$ graphs, form $2 N$ graphs $\{\tilde{\mathcal{G}}_i^1, \tilde{\mathcal{G}}_i^2\}$ through deliberate graph augmentations;
    \item Update text and graph encoder through gradient descent on a loss designed to promote proximity between matching cross-modality $(\mathcal{T}_i, \tilde{\mathcal{G}}_i)$ and graph $(\tilde{\mathcal{G}}_i^1, \tilde{\mathcal{G}}_i^2)$ embedding pairs from the same molecule, and higher distance between non-matching pairs.
\end{itemize}

We implement the InfoNCE loss function \cite{InfoNCE}, comprising of a term for graph pairs and a term for text-graph pairs (here the cross-modality pair):
$$ \ell (\mathcal{T}_i, \tilde{\mathcal{G}}_i) = - \textrm{log} \frac{\textrm{exp} \left( \textrm{cos} ( \textbf{z}^{\mathcal{T}}_i, \textbf{z}^{\mathcal{G}}_{i} ) / \tau  \right)}{\sum_{j \neq i}  \textrm{exp} \left( \textrm{cos} ( \textbf{z}^{\mathcal{T}}_i, \textbf{z}^{\mathcal{G}}_{j} ) / \tau \right)} $$

\subsubsection{Pre-trained text and graph encoders}

The goal of contrastive pre-training is to align the representations of matched text fragments and molecular 2D graphs in the same embeddings space. For efficiency purposes, we start with previously pre-trained models for both our text encoder and our graph encoder, which we present.

To optimize for extraction of information from fragment of scientific papers, we base our text encoder on SciBERT \cite{SciBERT}, a pre-trained language model based on BERT \cite{devlin2019bert}, trained on a large multi-domain corpus of scientific publications to improve performance on downstream scientific NLP tasks.

For our graph encoder, we use the GraphCL 80 pre-trained model \cite{you2020graph}, a 1.9 million parameters Graph Isomorphism Network (GIN) pre-trained through graph contrastive learning on \emph{MoleculeNet} \cite{moleculenet}.

\subsection{Relevance-Based Sampling}

\subsubsection{Neural Text Relevance Scoring} 

The \emph{MoMu} baseline retrieves text sequences by uniformly sampling two paragraphs associated with a molecule per epoch. As mentioned by the authors themselves, this approach assumes equal relevance of the retrieved paragraphs to the molecule's properties \cite{su2022}.

To address this issue, we propose a neural text retrieval strategy informed by the relevance of each text segment for the molecule it describes. For each paragraph, we compute the cosine similarity between the SciBERT CLS token embeddings for (i) the paragraph and (ii) a query:
\begin{itemize}
    \item \textbf{Mean similarity}: average embedding of the molecule name and its top 20 synonyms (see section \ref{sec:dataset})
    \item \textbf{Max similarity}: maximum similarity with any of the molecule name or its top 20 synonyms
    \item \textbf{Sentence similarity}: cosine similarity with a natural language query consisting of the following sentence:
\end{itemize}

\begin{quote}
 \emph{``Molecular, chemical, electrochemical, physical, quantum mechanical, biochemical, biological, medical and physiological properties, characteristics, and applications of \{NAME\}, a compound also known as \{SYNONYM$_1$\}, …, \{SYNONYM$_i$\}, …, or \{SYNONYM$_N$\}.''} 
\end{quote}

We then apply \textbf{epsilon sampling} \cite{hewitt2022truncation} to rank paragraphs by the cosine score and sample only from scores above a threshold, using the probability distribution (re-normalized over the strictly positive terms) and a temperature hyper-parameter to skew the sampling distribution towards the highest cosine score terms: 
$$ \mathbb{P}(\mathcal{T}_{i \in [1 .. N]}) =  \textrm{Softmax} \left( \frac{\textrm{cos} ( \textbf{z}_{query}, \textbf{z}_{i} ) }{ \textrm{Temp} } \right)   \quad \textrm{if } \geq \frac{\epsilon}{N}  $$

\subsubsection{Chemically-Valid Principled Graph Augmentations}

The baseline model is trained through the general contrastive learning strategy of modifying graphs by randomly dropping nodes and subgraphs, which had been shown to be effective for chemical tasks in the past \cite{you2020graph}. However, the resulting molecule graphs may not make physical sense. 

Here, we introduce a graph augmentations inspired by bio-chemical reactions, which lead to chemically valid augmented graphs. Specifically, we implement augmentations $\{\tilde{\mathcal{G}}_i^1, \tilde{\mathcal{G}}_i^2\}$ to the molecular graph  $\mathcal{G}_i$ which add or remove functional groups corresponding to the following methylation/de-methylation (replacing a hydrogen with a $CH_3$ group or vice versa), amination/de-amination reactions (replacing a hydrogen with an $NH_3$ or vice versa):
$$\ce{R-H + CH_4 <=> R-CH3 + H2} $$
$$\ce{R-H + NH_3 <=> R-NH2 + H2}$$

Notably, methylation and amination involve \textit{adding} nodes to the molecular graph, instead of node dropping and random-walk subgraphs which \textit{remove} nodes.

Each of these augmentations is performed on randomly selected nodes of the graph at batch sampling time (see appendix \ref{sec:augmentation}). Crucially, we carefully control for chemical validity of the reactions, and update the molecular graph tensor to comply with fundamental chemical rules such a bond valences and implicit hydrogens count.

\section{Experiments}

\subsection{Molecular Property Prediction}
\label{sec:dataset}

We measure the performance impact of our novel augmentations and pretraining strategy using the downstream task of molecular property prediction \cite{moleculenet}. From pre-training, we obtain two encoders that embed molecular graph and text descriptions within the same joint latent space: $f_G: \mathcal{G} \to \mathbf{z}_{{\mathcal{G}}} \in \mathcal{Z}$ and $f_T: \mathcal{T} \to \mathbf{z}_{{\mathcal{T}}}  \in \mathcal{Z}$. We fine-tune our graph encoder for classification tasks by adding a classifier MLP layer, which we adapt and fine-tune to each specific downstream task and dataset: 
$$ \textrm{MLP}_\textrm{CLASSIFIER}(\cdot) \circ f_G: \mathcal{G} \to \mathbf{\hat{y}}_{{\mathcal{G}}}$$

We pre-train on the molecular graph-text pairs dataset presented in figure \ref{fig:dataset}, constructed in \citet{su2022}, which comprises of 15,613 graph-document pairs, with 37 million paragraphs or 47.5 gigabytes of text ($\sim$3 megabytes per molecule) from scientific articles, presented in appendix \ref{sec:pretraining-dataset}. We then evaluate our models by fine-tuning them on biochemical classification tasks from \emph{MoleculeNet} \cite{moleculenet}, a multi-faceted set of benchmark tasks and reference datasets. Specifically, we use 7 datasets from \href{https://deepchemdata.s3-us-west-1.amazonaws.com/index.html}{DeepChem} and their associated classification tasks (BACE, BBBP, Clintox, MUV, SIDER, Tox21, and ToxCast), which are all detailed in appendix \ref{sec:finetuning-dataset}.

We fine-tune and evaluate the graph encoder on seven MoleculeNet datasets: BACE \cite{bace}, BBBP \cite{bbbp}, Tox21, ToxCast \cite{toxcast}, SIDER \cite{sider}, ClinTox \cite{clintox}, and MUV \cite{muv}. We use Area Under Receiver-Operator Curve (AUROC) to measure performance and evaluate for three random seeds and report the mean and standard deviation in Table \ref{fig:results}.

\begin{table*}[!t]
    \resizebox{1\textwidth}{!}{%
    \begin{tabular}{llllllll}
\textbf{Experiment}                                                        & \textbf{BACE}     & \textbf{BBBP}     & \textbf{Tox21}    & \textbf{ToxCast}  & \textbf{SIDER}    & \textbf{ClinTox}  & \textbf{MUV}      \\
\\ \hline \\
Graph only pre-training & 70            & 65.8         & 74            & 63.4           & 57.3           & 58            & 71.8        \\
\\ \hline \\ 
Baseline (\emph{MoMu})                                                                    & 70.31 ±3.67       & 68.04 ±1.67       & 74.6 ±0.68        & 63.27 ±0.53       & 59.39 ±0.51       & 61.09 ±1.1        & {\bf 75.66 ±0.55} \\
Baseline (pruned)                                                            & 71.14 ±1.93       & 67.86 ±2.1        & 74.77 ±0.37       & 62.71 ±1.3        & 59.31 ±0.72       & 61.17 ±1.39       & 75.18 ±1.06       \\
Baseline (relevant)                                                          & 72.13 ±0.47       & 68.73 ±2.21       & 74.85 ±0.3        & 62.47 ±0.66       & 60.05 ±0.7        & 59.99 ±1.73       & 74.47 ±0.95       \\

\\ \hline \\

Mean cosine similarity (best)                                                & 72.6 ±2.77        & 68.48 ±1.68       & 74.54 ±0.7        & 63.37 ±0.72       & 60.07 ±0.41       & 61.36 ±3.36       & 75.07 ±1.13       \\
Max cosine similarity (best)                                                 & {\bf 72.71 ±0.59} & 68.27 ±2.35       & 74.77 ±0.45       & {\bf 63.73 ±0.59} & 60.14 ±1.05       & {\bf 62.28 ±1.61} & 75.15 ±1.07       \\
Sentence cosine similarity (best)                                                & 72.05 ±0.52       & 68.11 ±2.5        & {\bf 74.94 ±0.79} & 63.6 ±0.29        & 59.84 ±0.24       & 61.47 ±2          & 74.61 ±0.27       \\
Principled graph augmentation                                                           & 71.45 ±2.24       & {\bf 69.23 ±0.93} & 74.31 ±0.36     & 62.61 ±0.49       & {\bf 61.33 ±0.69} & 58.97 ±2.22       & 75.03 ±1.52      \\ \\ 
\end{tabular}%
}
\caption{Results of our experiments: AUROC classifier task performance for multiple random seeds for each \emph{MoleculeNet} dataset, reported for each pre-training experiment and baseline model/dataset.}
\label{fig:results}
\end{table*}

\subsection{Experiments}

We use the following baselines for experimentation:

\begin{itemize}
    \item \textbf{\emph{MoMu}:} the original model presented in \citet{su2022} which samples text and graph augmentations with a uniform random distribution;
    \item \textbf{Naive text relevance:} as a non-neural control for the impact of relevance selection, we create a naive dataset with only sentences where molecule names and their synonyms appear explicitly;
    \item \textbf{Pruning:} to control for the impact of a smaller (thus potentially less noisy) data, we prune the dataset and keep only the first 256 characters of each paragraph;
    \item \textbf{Single modality pre-training:} as a baseline to measure performance gains from aligning graph representations with text, we also report performance of a GIN trained only on the graph modality.
\end{itemize}

We run the following experimental protocol and report results in table \ref{fig:results}: 

\paragraph{Cosine similarity pre-processing:} to speed up retrieval at train time, we pre-compute the cosine similarity scores for each paragraph in the dataset, with each of the query types in our experiments (\texttt{mean}, \texttt{max}, \texttt{sentence}).

\paragraph{Cosine similarity retrieval:} we ran experiments on the 3 cosine similarity query types, with hyper-parameters selected via an intrinsic evaluation based on hand-labeling of a small sub-set of text paragraphs, presented in appendix (table \ref{tab:intrinsic}). For pre-trained each model for 30 epochs (2 hours each on an A100 GPU).

\paragraph{Chemically-relevant graph augmentations:} lastly, we trained a comparison model trained on a uniform random text sampling strategy, but with chemically-relevant molecular graph augmentations, for a full 30 epochs run (2 hours on an A100 GPU).

\subsection{Results}

We report our experiment in table \ref{fig:results}, where we observe a consistent improvement on baseline performance for 6 of the 7 \emph{MoleculeNet} molecular property prediction datasets and associated classification tasks. 

Overall, using our strategy, the AUROC performance metric for \textbf{molecular property prediction improves by an average of +4.26\% across \emph{MoleculeNet} classification tasks} compared to molecular representations trained on the graph modality alone. We found that \texttt{max} and \texttt{sentence} cosine similarity tend to outperform random draw most consistently, followed by \texttt{mean} cosine similarity, and that our principled graph augmentations markedly improved the results on the BBBP and SIDER datasets.

The performance increase with regards to \emph{MoMu} and the baselines controlling for text pruning (relevance-based and length-based) are +1.54\%, +1.59\% and +1.49\% respectively. Of note: the naive relevance strategy only improves performance by +0.06\% vs. baseline, and pruning the paragraphs decreases performance by -0.05\%. We conclude that it is the molecular property knowledge extracted from scientific papers that improves graph representations through the multimodal contrastive training process. 




\section{Conclusion}

We demonstrated an improved strategy for multimodal contrastive learning of molecule representations from text corpora with principled augmentation and neural relevance scoring at sampling time. Our approach outperforms the baseline model (MoMu) for the downstream task of molecular property prediction on most \emph{MoleculeNet} datasets with an average performance gain of +1.54\%, and outperforms models trained on graphs only by +4.26\%.

Our results provide strong evidence that natural language encodes key knowledge on the properties of molecules. Extracting this information effectively through a deliberate alignment of graph representation and text embeddings is a powerful approach to improve property prediction models, and holds clear promise for computational biology and molecular sciences.

\section*{Acknowledgements}

The authors wish to thank Christopher Manning and John Hewitt for their instrumental feedback on neural relevance scoring, natural language sentence retrieval, and intrinsic evaluation; as well as Jure Leskovec and Mohammad Aljubran, on molecular 2D graphs representation and augmentation. We also wish to acknowledge \citet{su2022} for the original molecular multimodal contrastive learning implementation, \citet{you2020graph} for the pre-trained GraphCL GNN models, \citet{SciBERT} for the pre-trained SciBERT model, the Allen Institute for the S2ORC dataset \citep{s2orc}, and the National Institutes of Health for the \emph{PubChem} database \citep{PubChem}.

\bibliography{references}
\bibliographystyle{icml2023}

\newpage
\appendix
\onecolumn

\section{Appendix}

\begin{figure*}[!h]
    \centering
    \includegraphics[width=\textwidth]{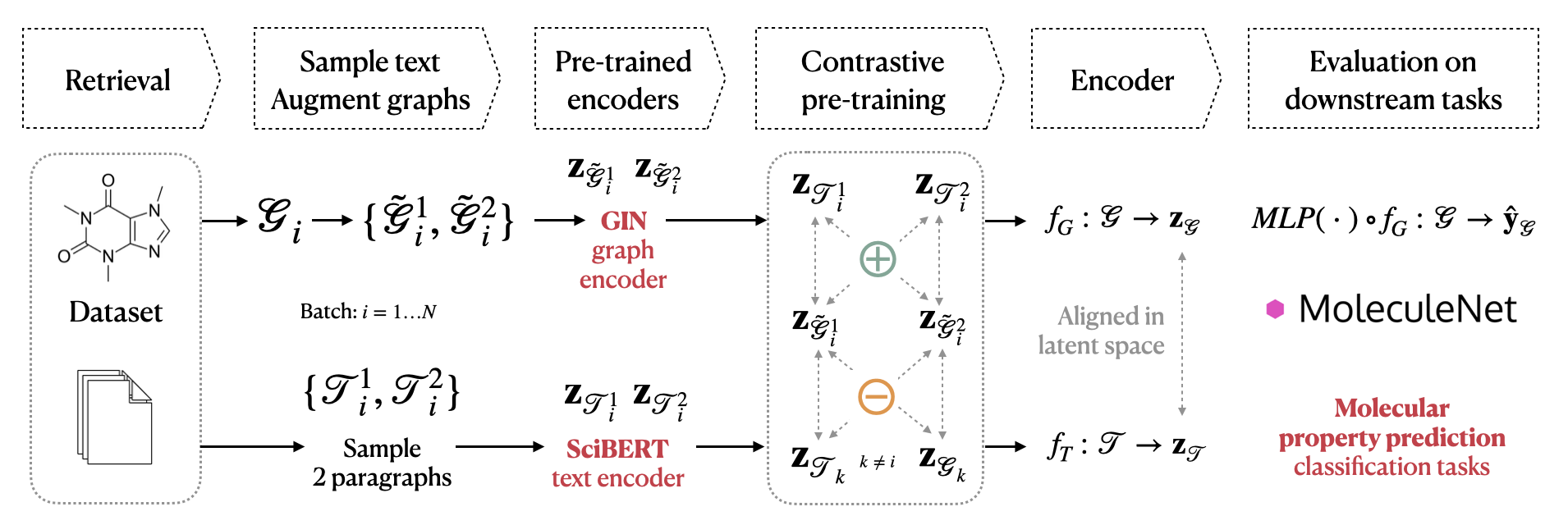}
    \caption{Contrastive pre-training of joint representations of molecular graph-text. Our contribution focuses on improvements to the \textbf{text retrieval and graph augmentation strategies}, which we evaluate on \textbf{downstream property prediction tasks}.}
    \label{fig:contrastive-learning}
\end{figure*}

\subsection{Pre-Training Dataset}
\label{sec:pretraining-dataset}

\begin{figure*}[!h]
    \centering
    \includegraphics[width=0.75\textwidth]{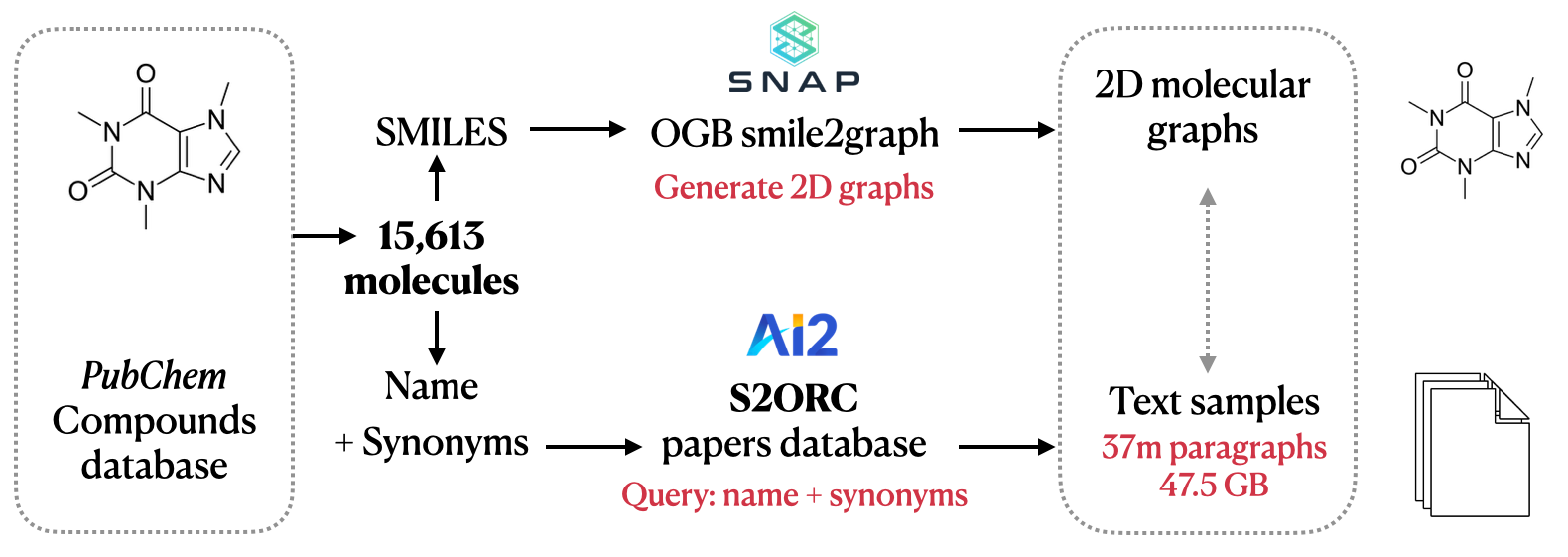}
    \caption{Joint molecular graph-text samples data set based on the \emph{PubChem} and S2ORC database.}
    \label{fig:dataset}
\end{figure*}

We train on the molecular graph-text pairs dataset presented in figure \ref{fig:dataset}, constructed following \cite{su2022} by retrieving scientific papers in the S2ORC \citep{s2orc} database by using the name and synonyms of compounds from \emph{PubChem} \citep{PubChem} as query, and transforming their SMILES intro a molecular graph using OGB smile2graph \cite{OGB}. 

The dataset comprises of 15,613 graph-document pairs, with 37 million paragraphs or 47.5 gigabytes of text ($\sim$3 megabytes per molecule). To make training tractable, the text beyond the first 500 paragraphs per molecule is left out. 

Importantly, the molecule graph and text sequences datasets are only weakly correlated: text fragments are extracted form the original SO2RC database on the basis of the name of the molecule appearing in that paragraph, with no further controls for relevance.
   
Lastly, the dataset is highly bi-modal: out of 15,613 text-graph pairs, 8,700 samples have less than 50 paragraphs of text, and 2,967 molecules have $\geq$500 paragraphs. Our sampling strategies based on cosine similarity scores aim to counter this inherent imbalance, by training on most of the small text corpus for the sparsely described molecules, and only relevant text for richly described ones.

\subsection{Downstream Molecular Property Prediction}
\label{sec:finetuning-dataset}

We use the following datasets retrieved from \href{https://deepchemdata.s3-us-west-1.amazonaws.com/index.html}{DeepChem}:

\begin{itemize}
    \item BACE: classification of inhibitors of a human enzyme involved in Alzheimer, which, if blocked, may prevent build up of proteins in the brain associated with the disease.
    \item BBBP: classification for the prediction of blood-brain barrier penetration by small molecules.
    \item Clintox: classification of drugs approved/rejected by the FDA for toxicity.
    \item MUV: classification for virtual molecule screening built on \emph{PubChem}.
    \item SIDER: classification of adverse side reactions of marketed drugs.
    \item Tox21: classification of toxicity measured by biological reactions and stress response.
    \item ToxCast: classification over 600 tasks linked to \emph{in vitro} toxicology data.
\end{itemize}



\subsection{Chemically-Valid Principled Graph Augmentations}

\label{sec:augmentation}

We implement algorithm \ref{alg:example}:

\begin{algorithm}[!h]
   \caption{Chemically-Valid Principled Graph Augmentations. \\ \textit{Example: methylation reaction, addition of a \ce{-CH3} functional group to the molecular group.}}
   \label{alg:example}
\begin{algorithmic}

\REQUIRE PyG graph tensor $x_i$, node features, edge features
\STATE \textbf{1. Randomly sample nodes} that are C atoms with implicit hydrogen count $\geq 1$
\STATE \textbf{2. Add a new node} to the graph for the additional functional group and update node features for valid covalence and implicit hydrogen numbers
\STATE \textbf{3. Add an edge} to the molecule graph with a single bond feature to bind the additional functional group
\STATE \textbf{4. Decrease implicit hydrogen count} for the original site to account for functional group addition

\end{algorithmic}
\end{algorithm}

\subsection{Intrinsic Evaluation for Hyper-parameter Search} 

To inform our search for the hyper-parameters with which to compute cosine similarity scores for sampling purposes, we ran an intrinsic evaluation of several potential retrieval methods and hyper-parameters. 

We hand labeled each paragraph in a small subset of text samples, and used paragraphs which all labelers classified as relevant to the molecule as the ground truth for our retrieval problem. 

We controlled for consistency between different human labelers by using Cohen's Kappa (\cite{kappa}). We report a score of 0.4874. 

We varied the temperature and epsilon hyper-parameters and computed recall, precision and F1 score based on the ground truth from hand labeling. Results for the mean similarity query schema are reported in figure \ref{tab:intrinsic}. 

On the basis of these results, we chose to run our cosine similarity pre-training experiments with $\epsilon = 0.5$ and Temperature $= \{ 0.05, 0.1, 0.2 \}$.

\begin{table*}[!h]
\centering
\begin{tabular}{lllllll}
\textbf{Temperature}      & \textbf{0.05}         &  \textbf{0.1}         & \textbf{0.2}        & \textbf{0.05}         &  \textbf{0.1}         & \textbf{0.2}        \\
\textbf{$\mathcal{E}$-threshold}    &  \textbf{0.5}   &  \textbf{0.5}   &  \textbf{0.5}     & \textbf{1}   & \textbf{1}   & \textbf{1} \\          
 \hline \hline \\
Recall    & 0.5          & 0.7419 & 0.9354  & 0.3          & 0.4375       & 0.5          \\
Precision & 0.5172 & 0.5227 & 0.5178 & 0.5294 & 0.56         & 0.5172 \\
F1 score  & \underline{0.5085} & \underline{0.6133} & \underline{0.6667} & 0.383 & 0.4912 & 0.5084 \\
\\ \hline \\
\end{tabular}
\caption{Intrinsic evaluation for the selection of epsilon sampling hyper-parameters.}
\label{tab:intrinsic}
\end{table*}

\subsection{Future work}
Evaluation of deep generative tasks in general, and molecular generation tasks in particular, is an open challenge in machine learning \citep{yousefzadegan2022evaluation}. As a next step, we could use the graph encoder and text encoder we trained to train generative models that help bridge these two modalities, with multi-model tasks such as:

\begin{itemize}
    \item Molecular captioning: given a molecular graph, generating text that accurately describes the molecule and its properties;
    \item Molecular generation: given a text description of desiged properties in natural language, generate a graph for a molecule that exhibits such properties.
\end{itemize}

For that purpose, and following \citet{su2022}, we implemented MoFlow \cite{MoFlow}, a flow-based deep generative model, on our pre-trained graph encoders, to experiment with molecular generation from free text. This showed very promising results for zero-shot molecular generation (zero-shot since we did not fine-tune the flow model to match out encoder specifically). A logical avenue for future work could be to use our trained graph encoder as a teacher model to train our own flow or diffusion-based model and measure improvements in molecular generation capacity.


\end{document}